\documentclass{fairmeta}
\usepackage[most]{tcolorbox}

\usepackage{amsmath}
\usepackage{amssymb}
\usepackage{enumitem}
\usepackage{wrapfig}
\usepackage{bm}

\title{\LARGE FLASH: Efficient Visuomotor Policy via Sparse Sampling}

\author[*]{Jiaqi Bai}
\author[*]{Jindou Jia}
\author{Yuxuan Hu}
\author{Gen Li}
\author{Xiangyu Chen}
\author{Tuo An}
\author{Kuangji Zuo}
\author[\dagger]{Jianfei Yang}

\affiliation{MARS Lab, Nanyang Technological University, Singapore}

\contribution[*]{Equal contribution. \\ \textsuperscript{$\dagger$}Corresponding author.}

\abstract{
Generative models such as diffusion and flow matching have become dominant paradigms for visuomotor policy learning, yet their reliance on iterative denoising incurs high inference latency incompatible with real-time robotic control. We present \textbf{F}ast \textbf{L}egendre-polynomial \textbf{A}ction policy via \textbf{S}parse \textbf{H}istory-anchored flow (\textbf{FLASH} Policy), which replaces discrete action-chunk generation with continuous \textit{Legendre} polynomial trajectory representation. Specifically, by fitting expert demonstrations under sparse temporal sampling, FLASH enables a single inference to cover a significantly extended action horizon. To further accelerate generation, FLASH initiates the flow matching process from history polynomial coefficients rather than uninformative \textit{Gaussian} noise, shortening the transport distance and enabling accurate single-step inference. Moreover, analytic polynomial differentiation directly provides desired velocity feed-forward signals to the torque controller without numerical approximation. Extensive experiments on five simulated and two real-world manipulation tasks demonstrate that FLASH achieves state-of-the-art success rates ($\ge 92\%$ across all tasks), a per-episode inference time of $31.40\,ms$ (up to $175\times$ faster than diffusion policies and $18\times$ faster than prior flow matching policies), up to $4\times$ faster training convergence than ACT, and $5\times$ to $7\times$ reduction in controller tracking error compared to discrete-action baselines.}

\correspondence{Jianfei Yang (\email{jianfei.yang@ntu.edu.sg}), Jiaqi Bai (\email{baij0018@e.ntu.edu.sg}) \\
\textbf{Project Page: }\url{https://b1ue-jay.github.io/FLASH}}

\begin{document}

\maketitle

\graphicspath{figure/}

\let\cite=\citep

\section{Introduction}

Imitation learning has enabled robots to acquire complex manipulation skills through expert demonstrations. Recently, generative modeling such as diffusion policy~\citep{chi2025diffusion} and flow matching~\citep{lipman2023flow} has emerged as a dominant paradigm for policy representation in imitation learning, exhibiting exceptional multi-modal distribution modeling capabilities. However, their reliance on iterative denoising or ordinary differential equation (ODE) solving leads to high inference latency, making it difficult to satisfy real-time control requirements in robotic systems.

To mitigate inference latency, prevailing methods primarily aim to reduce the number of inference steps~\citep{song2020denoising,song2023consistency,meng2023distillation,salimans2022progressive,sauer2024adversarial,geng2025mean,jia2026action}. 
Despite these advancements, they still rely on discrete action-point generation that requires high-frequency inference over short action chunks, where each chunk typically spans only a short duration (e.g., $\sim 0.8$ seconds in $\pi_0$~\citep{black2024pi_0}). A natural approach is to increase the prediction horizon to reduce inference frequency and improve real-time efficiency. However, doing so for a single prediction inevitably leads to a substantial expansion in output dimensionality, resulting in the ``curse of dimensionality'' and significantly increased computational cost. This fundamental trade-off raises a key question: \textbf{Can low-frequency inference meet high-frequency control demands without sacrificing control fidelity or increasing model complexity?}

\begin{figure}[ht]
\centering
\includegraphics[width=\linewidth]{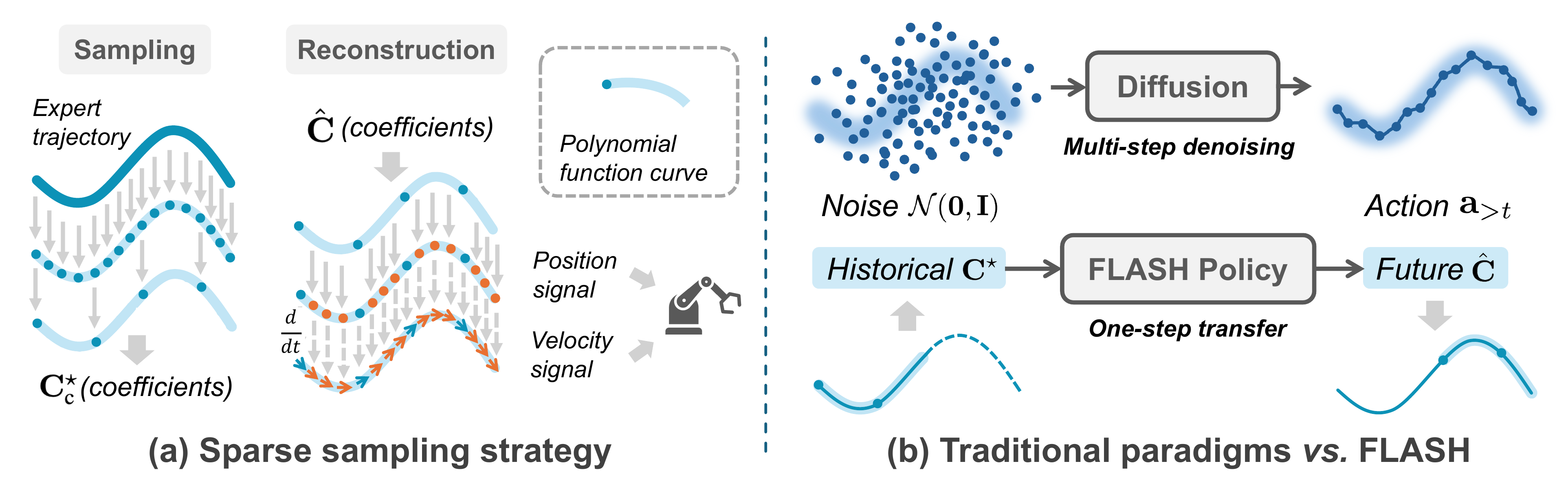}
\caption{\textbf{Overview of FLASH Policy.} \textbf{(a)}~\textbf{Sparse sampling strategy}: expert trajectories are fitted to polynomial coefficients $\mathbf{C}^{\star}_{\text{c}}$ at a significant long temporal stride. During deployment, we perform dense temporal upsampling on the predicted coefficients $\hat{\mathbf{C}}$. Both position and velocity signals are fed to the controller. \textbf{(b)}~\textbf{Traditional paradigms vs.\ FLASH}: conventional generative policies (top) generate discrete, non-smooth action points through multi-step denoising from uninformative noise. FLASH (bottom) operates in the coefficient space, transferring the historical coefficients to future coefficients with only one step, yielding a smooth continuous trajectory.}
\vspace{-17pt}
\label{fig:teaser}
\end{figure}

Our key observation is that robot motions are inherently smooth and low-frequency, and a simple \textit{Legendre} polynomial can represent the discrete action points with only a few coefficients. While there exist recent works adopting polynomial-based representations for trajectory planning~\citep{carvalho2025motion,singha2026crowd}, they still use the original number of action points to fit the polynomial. To achieve highly efficient inference, we propose \textbf{F}ast \textbf{L}egendre-polynomial \textbf{A}ction policy via \textbf{S}parse \textbf{H}istory-anchored flow (\textbf{FLASH}) that incorporates sparsity and generation mode for polynomial policy learning. Firstly, a \emph{sparse temporal sampling} strategy is proposed to fit polynomial coefficients at a significantly long temporal stride, as Fig.~\ref{fig:teaser} (a), so that each inference can cover a substantially extended action horizon without increasing model scale, directly answering the question above. Secondly, we introduce FLASH flow matching mechanism that initiates generation from the polynomial fitted to the measured action history, as Fig.~\ref{fig:teaser} (b). This keeps the entire generative process in a compact, structured coefficient space, dramatically shortening the transport distance and enabling accurate single-step inference. Moreover, to alleviate discontinuities between adjacent polynomial chunks, \textit{cross-horizon kinematic continuity} constraints are introduced to guarantee $C^1$ continuity (position and velocity)~\citep{abramowitz1948handbook} at the transition points.

The polynomial representation offers additional benefits for low-level controllers. First, its differentiable nature enables analytic computation of desired velocity via simple differentiation, which we leverage as a feedforward signal to the low-level torque controller to substantially reduce end-effector tracking errors. Second, this representation fundamentally guarantees the spatial smoothness of the trajectory and aligns with the varying control frequencies across different robotic platforms, which is hardware-friendly to low-level controllers. Third, during inference, task execution can be accelerated or decelerated by simply adjusting the evaluation temporal stride, satisfying the specific timing constraints of different tasks. We thoroughly evaluate FLASH in both simulated environments and real-world robotic systems. The empirical results show that FLASH exhibits superior performance regarding the success rate, efficiency, controller tracking error, and smoothness, consistently outperforming nine state-of-the-art (SOTA) baselines. 

To recap, the primary contributions of this work are as follows:
\begin{itemize}[leftmargin=0.5cm]
    \item We propose \textbf{F}ast \textbf{L}egendre-polynomial \textbf{A}ction policy via \textbf{S}parse \textbf{H}istory-anchored flow (\textbf{FLASH} Policy), which compresses a significantly long trajectory as only a few \textit{Legendre} polynomial coefficients fitted under sparse temporal sampling. Meantime, FLASH initiates the generative flow from a history prior in the coefficient space, enabling accurate single-step inference.
    \item We further introduce \textit{cross-horizon $C^1$ kinematic continuity} constraints for smooth chunk transitions, and leverage the differentiability of polynomials to feed the analytic velocity feed-forward signals to the torque controller without numerical approximation or additional training supervision.
    \item Extensive experiments across seven tasks demonstrate that FLASH achieves state-of-the-art success rates ($\ge92\%$ across all tasks), a per-episode inference time of $31.40\,ms$ (up to $175\times$ faster than diffusion policy~\citep{chi2025diffusion} and $18\times$ faster than prior flow matching~\citep{lipman2023flow}), up to $4\times$ faster training convergence than ACT~\citep{zhao2023learning}, and $5\times$ to $7\times$ controller tracking error reductions.
\end{itemize}

\section{Related work}

\subsection{Generative visuomotor policies}

Diffusion policy~\citep{chi2025diffusion} first formulates visuomotor control as conditional denoising diffusion process~\citep{ho2020denoising}, generating action chunks via iterative DDPM sampling.
Flow matching~\citep{lipman2023flow} offers a simulation-free alternative with more flexible ODE solvers. Subsequent efforts focus on accelerating diffusion and flow inference through faster samplers (DDIM~\citep{song2020denoising}, score matching~\citep{song2020score}), consistency models~\citep{song2023consistency}, diffusion distillation~\citep{salimans2022progressive,meng2023distillation,sauer2024adversarial}, MeanFlow~\citep{geng2025mean}, more expressive architectures (DiT~\citep{peebles2023scalable}), and visual tokenization with flow-based generation (VITA~\citep{gao2026vita}). While A2A~\citep{jia2026action} enables faster generation by replacing the uninformative \textit{Gaussian} prior with the previous \emph{action} chunk, it still operates in the discrete action space. Instead, FLASH operates in the \emph{functional} level. The generative flow is initiated from the polynomial coefficients fitted to the measured action history, a trajectory-level prior that preserves richer kinematic structure than a flat action-chunk prior. More broadly, all the above methods predict fixed-length discrete action chunks that inherently suffer from inter-chunk jitter and require high-frequency re-inference. In contrast, FLASH lifts the representation to continuous polynomial coefficients, decoupling inference frequency from control frequency.

\subsection{Polynomial trajectory representations}

Recent learning-based methods have begun to adopt polynomial and spline parameterizations in robot trajectory planning.
MPD~\citep{carvalho2025motion} trains diffusion models over B-spline control points, employing analytically derived velocity and acceleration solely as smoothness cost functions during the denoising process.
FlowMP~\citep{nguyen2025flowmp} extends this idea to flow matching, jointly learning position, velocity, and acceleration fields from B-spline training data, which triples the supervision dimensionality and increases training overhead. BEAST~\citep{zhou2025beast} employs B-splines purely as an efficient tokenization scheme for transformer-based imitation learning, without exploiting derivative information.
Crowd-FM~\citep{singha2026crowd} parameterizes trajectories as \textit{Bernstein} polynomials for mobile robot crowd navigation, yet does not feed the analytic velocity to any low-level controller.
A common limitation across these works is that \emph{none} actually leverages the analytic differentiability of the polynomial representation to supply velocity feedforward signals to a torque-level controller.
FLASH fills this gap. It learns inherently smooth \textit{Legendre}-coefficient trajectories with cross-horizon $C^1$ continuity, from which desired velocity is derived via analytical differentiation, without any additional training overhead.

\section{Fast legendre-polynomial action policy via sparse history-anchored flow}
\label{sec:method}

We consider a visuomotor imitation-learning setting in which a policy maps historical
observations to a future action trajectory. Let $\mathbf{a}_{\le t} = \{\mathbf{a}_{t-T_o+1}, \dots, \mathbf{a}_t\}$
denote the history actions~\footnote{We assume history actions are observable, a common setting in manipulation tasks where a direct semantic correspondence exists between actions and proprioceptive states (e.g.\ joint angle, end-effector pose, and their differential forms).} observed up to step $t$ with observation horizon $T_o$, $\mathbf{I}_{\le t}$ denote the corresponding
image stream, and $\mathbf{a}_{>t} = \{\mathbf{a}_{t+1}, \dots, \mathbf{a}_{t+T_a}\}$ denote the generated actions with action horizon $T_a$.
As illustrated in Fig.~\ref{fig:teaser}, FLASH departs from discrete action-chunk generation~\cite{zhao2023learning}: rather than learning $T_a$ discrete
action points, FLASH generates a small set of polynomial coefficients from which the entire continuous trajectory segment is analytically reconstructed. 
The key idea to improve inference efficiency is twofold. 
On the one hand, discrete action chunks are parameterized by compact polynomials (Sec.~\ref{sec:poly_param}), decoupling the network's output dimensionality from the prediction horizon. 
On the other hand, the generative flow is initiated from the polynomial coefficients fitted to the action history rather than from noise (Sec.~\ref{sec:p2p_flow}), shortening the transport distance and enabling accurate single-step generation. 
Sec.~\ref{sec:loss} details the learning objective.

\subsection{Sparse polynomial parameterization}
\label{sec:poly_param}

\paragraph{\textit{Legendre} trajectory representation.}
Instead of generating a sequence of discrete samples, we parameterize the action trajectory as a continuous function of a normalized time variable $s\in[0,1]$. Each action dimension is expanded on a set of orthogonal basis functions $\{\Phi_j\}_{j=0}^{K}$:
\begin{equation}
\label{eq:poly_expand}
\mathbf{a}(s) \;=\; \sum_{j=0}^{K} \mathbf{c}_{j}\,\Phi_{j}(s),
\qquad s = \frac{\tau - \tau_{\text{start}}}{\tau_{\text{end}} - \tau_{\text{start}}} \in [0,1],
\end{equation}
where $\Phi_{j}(s)\!\in\!\mathbb{R}$ is a scalar basis function, $\mathbf{c}_j \in \mathbb{R}^{d_a}$ is the corresponding coefficient vector, $K$ is the polynomial degree, $d_a$ is the action dimensionality, and $\tau_{\text{start}},\,\tau_{\text{end}}$ are the first and last timesteps of the segment. The polynomial spans $H_{\text{poly}}$ discrete steps; in its simplest form $H_{\text{poly}}=T_a$, covering the full action horizon. We adopt the shifted \textit{Legendre} polynomials~\citep{abramowitz1948handbook, szeg1939orthogonal} as $\{\Phi_j\}$, which are defined through the \textit{Bonnet} recurrence~\citep{spiegel1968mathematical} ($n=1,\dots,K\!-\!1$):
\begin{equation}
\label{eq:bonnet}
(n+1)P_{n+1}(x) = (2n+1)\,x\,P_{n}(x) - n\,P_{n-1}(x),
\qquad P_0(x)=1,\; P_1(x)=x.
\end{equation}
Evaluated at $x = 2s-1$, the \textit{Legendre} family is orthogonal on $[-1,1]$. This orthogonality ensures a well-conditioned \textit{Gram} matrix during data fitting and produces coefficient distributions with comparable magnitudes, which is critical for stable flow matching. With this continuous representation in place, we next describe how to fit the coefficients to expert demonstrations efficiently.

\paragraph{Sparse temporal sampling fitting.}
To extend the physical prediction horizon without increasing the network's output dimensionality, we apply a sparse temporal sampling strategy to the expert demonstrations. We extract nodes from the expert trajectory using a temporal stride $k$, so that $H_{\text{poly}}$ sparse fitting nodes effectively cover a physical time span of $k \times H_{\text{poly}}$ steps. Let $\mathbf{Q} \in \mathbb{R}^{H_{\text{poly}} \times d_a}$ be the stacked matrix of observed sparse actions, and $\mathbf{S} \in \mathbb{R}^{H_{\text{poly}} \times (K+1)}$ be the \textit{Legendre} basis matrix evaluated at the corresponding discrete timesteps. The optimal polynomial coefficients $\mathbf{C}^{\star}$ can be obtained in closed-form via ordinary least squares~\citep{boyd2004convex, weisberg2005applied} (OLS):
\begin{equation}
\label{eq:ols_fit}
\mathbf{C}^{\star} = (\mathbf{S}^{\top}\mathbf{S})^{-1}\mathbf{S}^{\top}\mathbf{Q}.
\end{equation}
This strategy leverages the analytic continuity of polynomials to compactly represent a trajectory segment spanning a longer physical horizon.

\paragraph{Cross-horizon continuity and two-tier extension.}
Solving $\mathbf{C}^{\star}$ independently for adjacent sub-segments introduces position or velocity discontinuities at the junctions. To enforce $C^1$ continuity, we extend the fitting window with overlapping regions on both sides of the execution interval and place exact kinematic anchors at the segment boundaries. A closed-form \textit{KKT} correction~\citep{mellinger2011minimum} then projects the unconstrained $\mathbf{C}^{\star}$ onto the constraint manifold, yielding the corrected coefficients $\mathbf{C}^{\star}_{\text{c}}$. We further append fit-padding steps to pull these anchors into the stable interior of the \textit{Legendre} fitting window, preventing boundary sensitivity. The corrected coefficients are then row-wise scale-normalized to stabilize the flow-matching objective. Detailed derivations are provided in Appx.~\ref{appendix:continuity}. The above pipeline produces the training targets. We now describe how the predicted coefficients are decoded into trajectories at deployment.

\paragraph{Temporal upsampling and frequency decoupling.}
Since the policy learns a trajectory representation under sparse sampling, its single inference output spans the physical time of $k \times H_{\text{poly}}$ steps window. During deployment, we perform dense temporal upsampling on the predicted coefficients $\hat{\mathbf{C}}$, evaluating the executable trajectory $\hat{\mathbf{q}}(s_i)$ and its exact higher-order kinematic derivatives either at the original frequency used to collect the expert trajectories, or at any arbitrary high-frequency grid required by different low-level controllers:
\begin{equation}
\label{eq:decode}
\begin{aligned}
\hat{\mathbf{a}}(s_i) &= \sum_{j=0}^{K} \hat{\mathbf{c}}_j \Phi_j(s_i),
&\quad
\hat{\mathbf{v}}(s_i) &= \frac{1}{T}\sum_{j=0}^{K}\hat{\mathbf{c}}_j \Phi'_j(s_i),
\end{aligned}
\end{equation}

where $T$ denotes the physical time span of the polynomial segment,
$T = k\,H_{\text{poly}}/f_{\text{expert}}$, with $f_{\text{expert}}$ the
sampling frequency of the original expert trajectories. This sparse sampling strategy drastically mitigates inference latency (by $1/k$) while using extended chunk lengths to smooth transitions at splicing boundaries. Moreover, the continuous polynomial representation
naturally supports post-hoc temporal scaling: a single inference
produces coefficients $\hat{\mathbf{C}}$ that fix the trajectory shape
on the normalized time $s\!\in\![0,1]$, while the physical duration $T$
over which this shape is played satisfies $T \propto k$
(Eq.~\eqref{eq:decode}); the velocity command $\hat{\mathbf{v}}\!\propto\!1/T$
is automatically rescaled by the same analytic derivative. Hence,
simply setting the evaluation stride $k_{\text{eval}}$ to a value
different from the training stride $k_{\text{train}}$ allows the same
generated polynomial to be executed at any desired speed, with no
retraining and no extra computation. This enables online acceleration
of simple, safe sub-tasks for shorter execution time and online
deceleration of high-precision, high-risk maneuvers. The achievable
speed range is empirically validated in Sec.~\ref{sec:speed_mod}.

\subsection{History-anchored flow}
\label{sec:p2p_flow}

With the polynomial representation, we employ a flow-matching formulation to predict the transition of coefficients from adjacent history to the future timesteps, conditioned on the observations.

\paragraph{Flow matching.}
Flow matching~\citep{lipman2023flow} learns a time-dependent velocity field $f_\theta$ that
transports a source distribution $p_0$ to a target distribution $p_1$ along the probability
path induced by the linear interpolant $\mathbf{a}_\tau = (1-\tau)\mathbf{a}_0 + \tau\mathbf{a}_1$,
$\tau\!\in\![0,1]$. The corresponding conditional velocity is
$\mathbf{v}^{\star} = \mathbf{a}_1 - \mathbf{a}_0$, and training minimizes
$\mathbb{E}_{\tau,\mathbf{a}_0,\mathbf{a}_1}\|f_\theta(\mathbf{a}_\tau,\tau,\mathbf{e}) - \mathbf{v}^{\star}\|^{2}$, where $\tau$ is the flow time uniformly sampled from $[0,1]$ and $\mathbf{e}$ denotes the conditioning observation embedding. Standard flow matching policies fix $p_0 = \mathcal{N}(\mathbf{0},\mathbf{I})$, which obliges the network to learn long transport
paths and motivates the many-step ODE integration used at inference.

\paragraph{History-anchored flow.}
In visuomotor control, consecutive action segments share strong kinematic structure~\citep{jia2026action}. We leverage this informative prior by defining the generative flow directly between history and future polynomials, rather than originating from uninformative \textit{Gaussian} noise. Specifically, we fit the observed proprioceptive history $\mathbf{a}_{\le t}$ to the \textit{Legendre} basis to obtain the history coefficients $\mathbf{C}_h$. We use the same scheme as Eq.~\eqref{eq:ols_fit}, but introduce \textit{Tikhonov} regularization $\lambda_{h}\!>\!0$. This formulation not only allows for fitting when the history window is configured to be short, but crucially, it mitigates the numerical ill-conditioning caused by dense discrete sampling over a narrow time horizon. By preventing coefficient explosion, it ensures a numerically stable and bounded starting distribution for the subsequent flow matching:
\begin{equation}
\label{eq:hist_fit}
\mathbf{C}_{h}
\;=\;
\big(\mathbf{S}_{h}^{\!\top}\mathbf{S}_{h} + \lambda_{h}\mathbf{I}\big)^{-1}
\mathbf{S}_{h}^{\!\top}\,\mathbf{a}_{\le t}.
\end{equation}
FLASH then constructs a straight-line flow path from $\mathbf{C}_{h}$ to the target future coefficients $\mathbf{C}_{1}$:
\begin{equation}
\label{eq:p2p_interp}
\mathbf{C}_\tau \;=\; (1-\tau)\,\mathbf{C}_{h} + \tau\,\mathbf{C}_{1},
\qquad
\mathbf{v}^{\star}(\mathbf{C}_\tau,\tau) \;=\; \mathbf{C}_{1} - \mathbf{C}_{h}.
\end{equation}

\begin{figure}
  \centering
  \vspace{-14pt}
  \includegraphics[width=\linewidth]{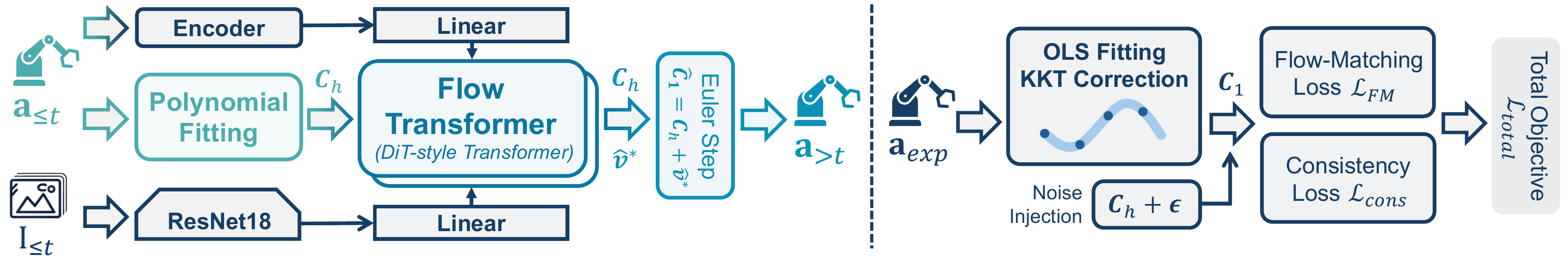}
  \caption{\textbf{Overview of FLASH's pipeline.} \textbf{Left: Inference.} History actions are fitted into Legendre-polynomial coefficients and fused with visual features by a DiT-style Flow Transformer. A single Euler step predicts future coefficients, which are decoded into executable actions. \textbf{Right: Training.} Expert actions are converted into target polynomial coefficients via OLS fitting and KKT correction, and the model is optimized with flow-matching and consistency losses.}
  \label{fig:p2p_flow}
  \vspace{-10pt}
\end{figure}

A DiT-style transformer~\citep{peebles2023scalable} $f_\theta(\mathbf{C}\tau,\tau,\mathbf{e})$ regresses this constant velocity $\mathbf{v}^{\star}$, where the global condition $\mathbf{e}$ is obtained by encoding the multi-view image stream $\mathbf{I}_{\le t}$ with a ResNet-18 encoder~\citep{he2016deep}, concatenating each frame's visual features with the corresponding proprioceptive history $\mathbf{a}_{\le t}$, and flattening across the $T_o$ observation steps, as shown in Fig.~\ref{fig:p2p_flow}. Because $\mathbf{C}_{h}$ is typically close to $\mathbf{C}_{1}$, the
transport distance is short and a single \textit{Euler} step $\hat{\mathbf{C}}_{1} = \mathbf{C}_{h} + f_\theta(\mathbf{C}_{h},0,\mathbf{e})$
is sufficient at inference. The resulting $\hat{\mathbf{C}}_{1}$ is then decoded into the executable action trajectory $\hat{\mathbf{a}}{>t}$ together with its analytical velocity profile via Eq.~\eqref{eq:decode}. During training, we inject minimal \textit{Gaussian} noise $\boldsymbol{\epsilon}\sim\mathcal{N}(\mathbf{0},\sigma_{h}^{2}\mathbf{I})$ into $\mathbf{C}_{h}$ to improve robustness against off-policy histories at deployment. With the parameterization (Sec.~\eqref{sec:poly_param}) and flow architecture defined, we now formalize the learning objectives.

\subsection{Learning objectives}
\label{sec:loss}

\paragraph{Flow-matching loss.}
Following Eq.~\eqref{eq:p2p_interp}, the network is trained to regress the constant transport
velocity along every intermediate point of the polynomial flow,
\begin{equation}
\label{eq:loss_fm}
\mathcal{L}_{\text{FM}}
\;=\;
\mathbb{E}_{\tau\sim\mathcal{U}[0,1],\,\mathbf{C}_{h},\,\mathbf{C}_{1}}
\Big\lVert f_\theta\!\big(\mathbf{C}_{\tau},\tau,\mathbf{e}\big) - (\mathbf{C}_{1} - \mathbf{C}_{h})\Big\rVert_{2}^{2}.
\end{equation}

\paragraph{Polynomial consistency loss.}
Eq.~\eqref{eq:loss_fm} supervises the vector field at arbitrary $\tau$, but does not
explicitly constrain the \emph{one-step} \textit{Euler} update used at inference. We therefore add a
consistency term that enforces transport of $\mathbf{C}_{h}$ to $\mathbf{C}_{1}$ in a single
evaluation of $f_\theta$ at $\tau=0$:
\begin{equation}
\label{eq:loss_cons}
\mathcal{L}_{\text{cons}}
\;=\;
\mathbb{E}_{\mathbf{C}_{h},\,\mathbf{C}_{1}}
\Big\lVert \mathbf{C}_{h} + f_\theta\!\big(\mathbf{C}_{h},\,0,\,\mathbf{e}\big) - \mathbf{C}_{1}\Big\rVert_{2}^{2}.
\end{equation}
This term bridges the gap between the continuous-time flow-matching objective and the
discrete one-step ODE solver used at deployment, and is indispensable for retaining accuracy
in the $N_{\text{NFE}}=1$ regime (See ablation study in Sec.~\ref{sec:ablation}).

\paragraph{Total objective.}
With a scalar weight $\lambda_{\text{cons}}>0$, the training loss reads
\begin{equation}
\label{eq:loss_total}
\mathcal{L}_{\text{total}} \;=\; \mathcal{L}_{\text{FM}} \;+\; \lambda_{\text{cons}}\,\mathcal{L}_{\text{cons}}.
\end{equation}
All parameters of the vision encoder, condition projector, and flow transformer
$f_\theta$ are optimized jointly. The scale-normalization matrix, least-squares solver (Eq.~\eqref{eq:ols_fit}), \textit{Tikhonov}-regularized history solver (Eq.~\eqref{eq:hist_fit}), and \textit{KKT} correction (Eq.~\eqref{eq:kkt}) are all precomputed and remain fixed during training.

\section{Experiments}
\label{sec:eval}

\begin{figure}[ht]
\centering
\includegraphics[width=\linewidth]{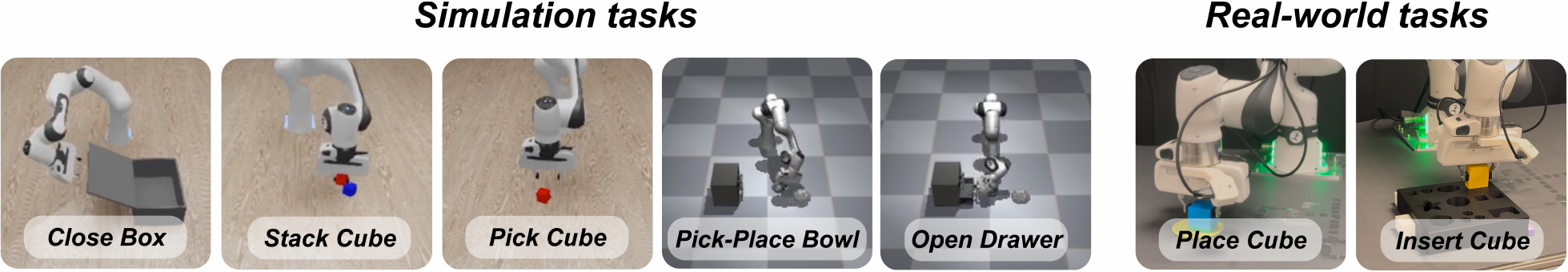}
\caption{\textbf{Left:} Five simulated tasks from \textit{Roboverse}~\cite{geng2025roboverse} platform. \textbf{Right:} Two real-world tasks.}
\label{fig:tasks}
\end{figure}

To evaluate the proposed method, we conduct the experiments in both simulated and real-world environments. 
As illustrated in Fig.~\ref{fig:tasks}, the evaluation encompasses seven manipulation tasks on \textit{Franka} robot, five of which are in simulation on \textit{Roboverse}~\citep{geng2025roboverse} platform, and two are real-world tasks. 

For systematical comparison, we select nine state-of-the-art baselines: DDPM-UNet~\citep{chi2025diffusion,ho2020denoising}, DDPM-DiT~\citep{ho2020denoising,peebles2023scalable}, DDIM-UNet~\citep{chi2025diffusion,song2020denoising}, FM-UNet~\citep{lipman2023flow}, FM-DiT~\citep{lipman2023flow,peebles2023scalable}, Score-UNet~\citep{song2020score}, ACT~\citep{zhao2023learning}, VITA~\citep{gao2026vita}, A2A-Noise~\citep{jia2026action}, and FLASH-Gaussian (FLASH-G). 
Specifically, FLASH-G is developed to serve as our homologous control group. It perfectly shares the network architecture for polynomials, \textit{Legendre} basis and sparse sampling training strategy with FLASH. The only difference is that FLASH-G initiates flow matching from pure \textit{Gaussian} noise ($p_0=\mathcal{N}(0,\mathbf{I})$), representing the standard generation from noise paradigm similar to traditional generative paradigm. Therefore, the performance gap between FLASH and FLASH-G strictly isolates the independent contribution of the proposed ``history-anchored flow'' mechanism. Task descriptions, demonstration counts, and all hyperparameters are detailed in Appx.~\ref{appendix:eval_setup}.

\begin{table*}[h]
\centering
\small
\setlength{\tabcolsep}{3.5pt}
\caption{Success rates (\%) on five tasks at a shared training budget of
$10k$ optimizer steps. \textbf{NFE} denotes the number
of generator function sampling steps at inference. Best results are \textbf{bold} while second-best results are \underline{underlined}.}
\label{tab:sr_main}
\begin{tabular}{lcccccc}
\toprule
\textbf{Method} & \textbf{NFE} & \textbf{Close Box} & \textbf{Pick Cube} & \textbf{Stack Cube} & \textbf{Open Drawer} & \textbf{Pick-Place Bowl} \\
\midrule
Score-UNet          & 100 & 44 & 58 & 24 & 2 & 8 \\ 
DDPM-UNet           & 100 & 84 & 68 & 44 & 76 & 90 \\ 
DDIM-UNet           & 40 & 80 & 74 & 42 & 76 & \underline{92} \\
DDPM-DiT            & 100 & 68 & 78 & 36 & 50 & 76 \\ 
FM-DiT              & 10 & 70 & 92 & 46 & 36 & 68 \\ 
FM-UNet             & 10 & 78 & 82 & 30 & 70 & 76 \\ 
ACT                 & 1  & 70 & \textbf{98} & 20 & 80 & 0 \\ 
VITA                & 6  & 94 & 86 & 86 & 68 & 86 \\ 
A2A-Noise                 & 1  & \underline{98} & 90 & \underline{92} & \underline{88} & \underline{92} \\ 
FLASH-G                & 10 & 70 & 94 & 82 & 62 & 88 \\ 
\textbf{FLASH (ours)} & 1  & \textbf{100} & \textbf{98} & \textbf{96} & \textbf{92} & \textbf{98} \\ 
\bottomrule
\end{tabular}
\end{table*}

As shown in Tab.~\ref{tab:sr_main}, FLASH achieves $\ge 92\%$ success rate across all five tasks with single-step inference (NFE=1), outperforming the homologous FLASH-G by 17.6\,pp on average (isolating the FLASH mechanism's contribution) and the strongest single-step baseline A2A-Noise by 4.8\,pp on average. Each entry is computed over 50 independent rollouts; at the median success rate of 92\%, the 95\% Clopper-Pearson confidence interval half-width is $\pm$7.5\,pp. Details are provided in Appx.~\ref{appendix:sr_analysis}.

Next, we provide a detailed statistical comparison from three perspectives: training efficiency (Sec.~\ref{sec:eval_training}), inference speed (Sec.~\ref{sec:eval_speed}), and controller tracking error (Sec.~\ref{sec:eval_smooth_tracking}). We also explore the possibility of adjusting the task execution speed through post-hoc temporal scaling (Sec.~\ref{sec:speed_mod}).

\subsection{Training efficiency}
\label{sec:eval_training}

\begin{figure}[h]
    \centering    \includegraphics[width=\textwidth]{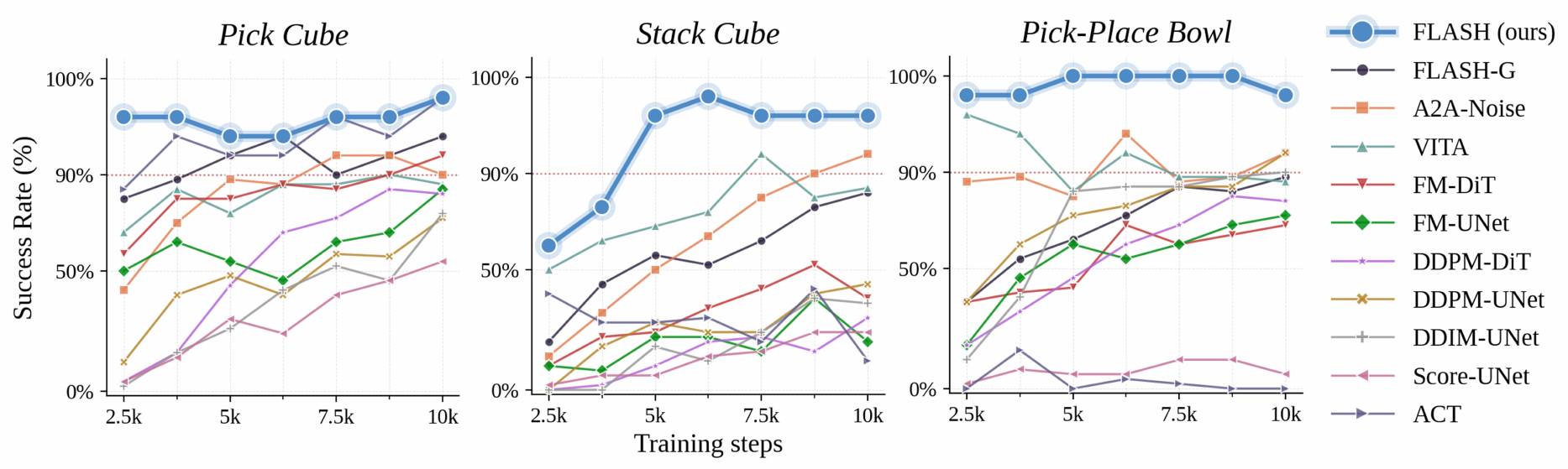} 
\caption{\textbf{Training efficiency.} Success rate versus training steps of 11 policies for three representative tasks.}
\label{fig:training_efficiency}
\end{figure}

Beyond a higher performance ceiling, FLASH exhibits significantly faster convergence. We evaluated the 11 policies across seven checkpoints at $\{2.5, 3.75, 5, 6.25, 7.5, 8.75, 10\} \times 10^3$ training steps under identical configurations. Fig.~\ref{fig:training_efficiency} illustrates the convergence speed on three representative tasks: \textit{Pick Cube}, \textit{Stack Cube}, and \textit{Pick-Place Bowl} (results for other tasks are provided in Appx.~\ref{appendix:training_efficiency}).

It can be seen in \textit{Pick Cube} task that FLASH reaches a 96\% success rate in just 2,500 steps, whereas the strongest baseline, ACT~\citep{zhao2023learning}, requires 10,000 steps to reach a comparable level, making FLASH approximately $4\times$ faster to train. In \textit{Stack Cube} task, FLASH achieves 98\% success at 6,250 steps, while the most competitive baseline at that point (VITA~\citep{gao2026vita}) sits at only 72\%, a gap of nearly 30 percentage points. In \textit{Pick-Place Bowl} task, FLASH stabilizes at a perfect 100\% success rate after 5,000 steps, consistently outperforming all baselines throughout the remainder of the training. These rapid convergence properties demonstrate that operating in the history-anchored polynomial space drastically reduces the sample complexity and optimization burden compared to standard noise-to-action or discrete-action paradigms, allowing the policy to rapidly acquire robust behaviors.


\begin{figure}[ht]
\centering
\includegraphics[width=\linewidth]{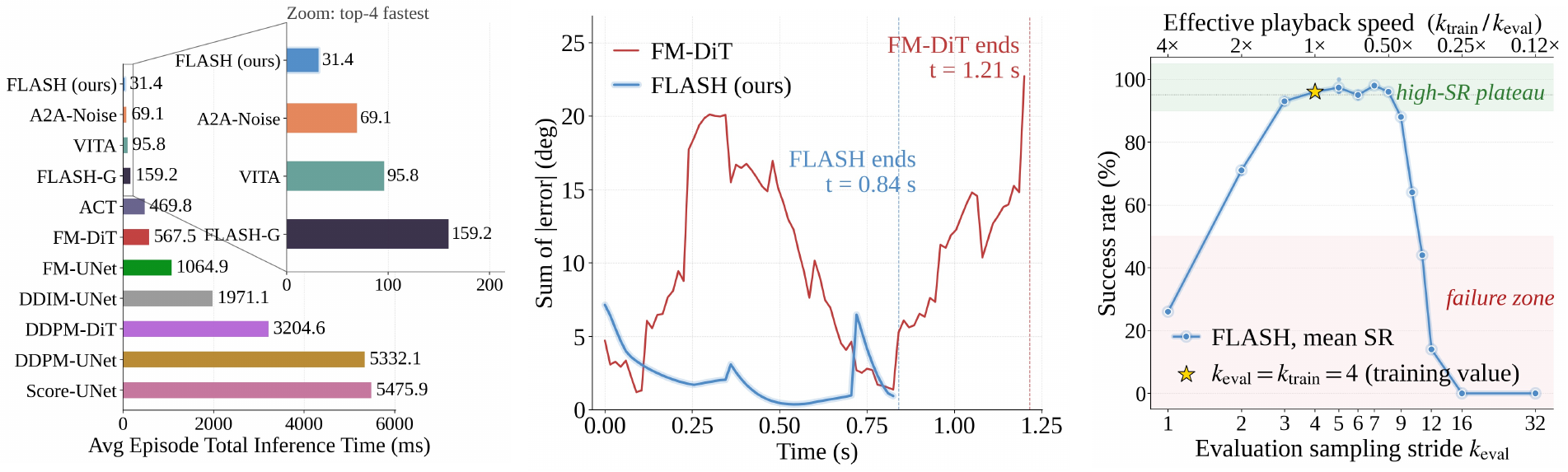}
\caption{\textbf{Simulation experiments.} \textbf{Left:} total episode inference time on \textit{Stack Cube} (all 11 policies, RTX 5090); inset zooms the top-4 fastest. \textbf{Middle:} sum of 7-joint absolute tracking errors on \textit{Pick Cube}; dashed lines mark task completion times. \textbf{Right:} post-hoc speed modulation on \textit{Stack Cube} by sweeping $k_{\text{eval}}$.}
\vspace{-10pt}
\label{fig:sim_exp}
\end{figure}

\subsection{Inference speed}
\label{sec:eval_speed}

Inference speed is evaluated across three progressive dimensions: (a) total episode inference time, (b) per-call latency, and (c) real-world task completion wall time. For fair comparison, all the selected policies are employed on the same PC with NVIDIA RTX 5090 GPU. In this part, we mainly display the ablation results on (a), and the results on (b) and (c) are detailed in Appx.~\ref{appendix:inference_speed}.


As shown in Fig.~\ref{fig:sim_exp} (left), this metric accumulates the duration of all inference calls within a successful rollout, directly reflecting the total computational cost required to complete a task. FLASH ranks first at $31.4\ ms$ (averaged over all successful episodes): $175\times$ faster than Score-UNet ($5476\ ms$), a difference of over two orders of magnitude; $2.2\times$ and $5.1\times$ faster than the strongest A2A-Noise (69~ms) and the homologous FLASH-G ($159\ ms$).
FLASH's speed advantage stems from two independent mechanisms: (i)~sparse sampling training (with a temporal stride $k=4$), where each polynomial covers $4\times$ the action horizon of discrete policies, lowering the number of inference calls per episode; and (ii)~the polynomial-to-polynomial flow starting from an informative history prior rather than noise, enabling single-step generation (NFE=1). The $5.1\times$ episode speedup over FLASH-G (which shares sparse sampling but starts from noise) isolates the contribution of this mechanism. See rigorous pairwise decomposition in Appx.~\ref{appendix:inference_speed}.


\subsection{Controller tracking error}
\label{sec:eval_smooth_tracking}

The polynomial representation in FLASH also benefits the low-level controller through analytic velocity feed-forward (Eq.~\ref{eq:decode}) and native high-frequency signal generation. We validate this on both simulation and real-world tasks.

\begin{figure}[ht]
\vspace{-6pt}
\centering
\includegraphics[width=0.70\linewidth]{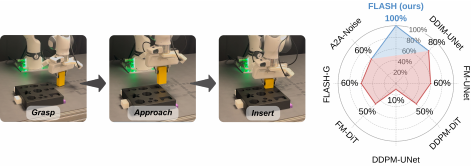}
\caption{\textbf{Real-world \textit{Insert Cube} task.} Execution snapshots (grasp $\rightarrow$ approach $\rightarrow$ insert) and success-rate radar chart across 8 policies under millimeter-level insertion tolerance.}
\vspace{-10pt}
\label{fig:insert_cube}
\end{figure}

In simulation (Fig.~\ref{fig:sim_exp}, middle), FLASH's mean and peak tracking errors are $4.70\times$ and $3.18\times$ lower than FM-DiT, whose error curve spikes at every chunk boundary. On the real-world \textit{Insert Cube} task with millimeter-level tolerance (Fig.~\ref{fig:insert_cube}), FLASH achieves 100\% success rate, leading the other 7 baselines by an average of 47\%. These results, together with the real-world tracking MAE analysis ($0.274\!\pm\!0.004^{\circ}$ for FLASH vs.\ $0.460\!\pm\!0.028^{\circ}$ for FM-DiT over 5 rollouts in Appx.~\ref{appendix:tracking_details}, confirm that control-level precision is jointly improved by three advantages of polynomial parameterization: trajectory smoothness, native high-frequency control output, and analytic velocity feed-forward. Detailed analysis is provided in Appx.~\ref{appendix:tracking_details}.

\subsection{Post-hoc execution speed modulation}
\label{sec:speed_mod}

Since the polynomial fixes the trajectory shape on a normalized time axis, execution speed can be modulated post-hoc by simply varying the evaluation stride $k_{\text{eval}}$ relative to $k_{\text{train}}$, with no retraining. As shown in Fig.~\ref{fig:sim_exp} (right), sweeping $k_{\text{eval}}$ on \textit{Stack Cube} reveals a wide high-success-rate plateau ($\ge$93\%) spanning $0.5\times$ to $1.33\times$ playback speed. A detailed analysis of the failure modes at extreme speeds is provided in Appx.~\ref{appendix:speed_mod}.

\section{Ablation study}
\label{sec:ablation}
To evaluate the contribution of each component, we conduct an ablation study on four key design choices in FLASH.
Full experimental details and extended analysis for each ablation are provided in Appx.~\ref{appendix:ablation}.


\begin{figure}[h]
\vspace{-1pt}
\centering
\includegraphics[width=\linewidth]{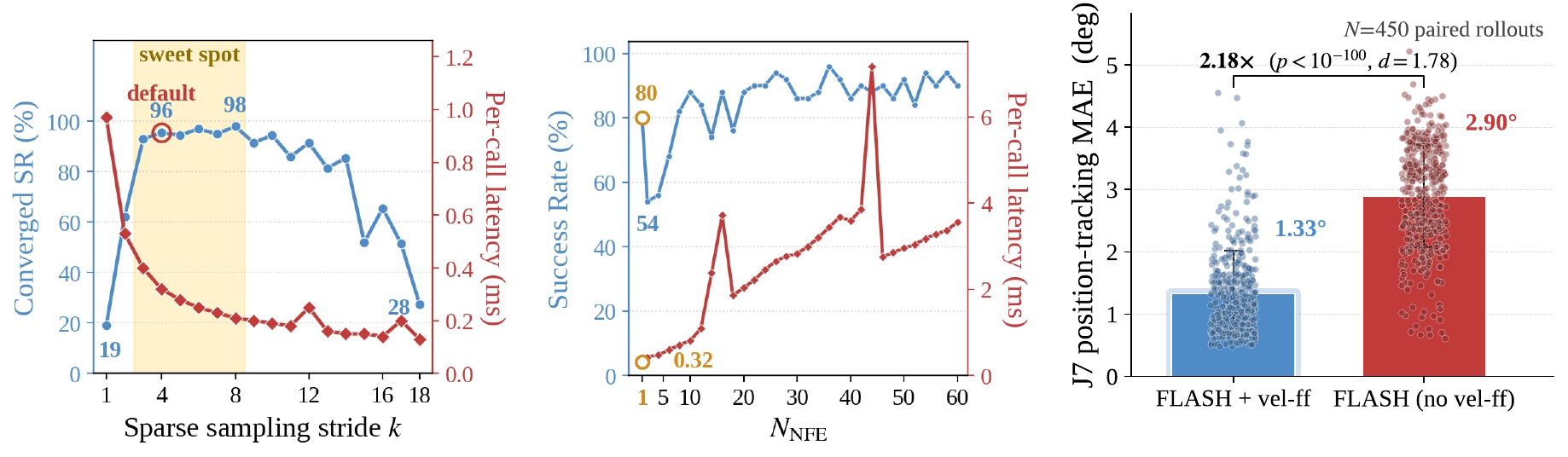}
\caption{\textbf{Ablation studies.} \textbf{Left:} effect of sparse sampling stride $k$ on \textit{Stack Cube}; yellow band marks the sweet spot $k\!\in\![3,8]$. \textbf{Middle:} $N_{\text{NFE}}$ sweep on \textit{Close Box}; latency (red) grows linearly. \textbf{Right:} velocity feed-forward ablation.}
\vspace{-3pt}
\label{fig:ablation}
\end{figure}

\textbf{Sparse sampling stride $k$.} Sweeping $k$ from $1$ to $18$ on \textit{Stack Cube} (Fig.~\ref{fig:ablation}, left) reveals that sparse sampling is essential: $k\!=\!1$ caps at $24\%$ success, while $k\!=\!4$ reaches $96\%$ ($+72$\,pp). A sweet spot at $k\!\in\![3,8]$ yields $\ge$93\% with monotonically decreasing latency; beyond $k\!\ge\!13$ performance collapses due to loss of polynomial expressivity and insufficient visual feedback.

\textbf{Fit padding and cross-horizon continuity. }On the \textit{Close Box} task, removing both fit padding~(Eq.~\eqref{eq:fitpad}) and KKT continuity~(Eq.~\eqref{eq:kkt}) yields only 75.0\%. Adding fit padding alone recovers 86.0\% ($+$11.0\,pp); further stacking the KKT $C^1$ correction (Eq.~\eqref{eq:kkt}) reaches 97.5\% ($+$11.5\,pp). Both mechanisms are necessary for the polynomial pipeline (see Tab.~\ref{tab:abl_fitpad} in Appx.~\ref{appendix:abl_fitpad} for details).

\textbf{Number of function evaluations $N_{\text{NFE}}.$} Single-step Euler ($N_{\text{NFE}}\!=\!1$) achieves $80\%$ success, \emph{outperforming} multi-step integration at $N_{\text{NFE}}\!\in\!\{2,4,6\}$ (Fig.~\ref{fig:ablation}, middle). This validates the consistency loss $\mathcal{L}_{\text{cons}}$ (Eq.~\eqref{eq:loss_cons}), which specializes the model for single-step generation. Since latency grows linearly with $N_{\text{NFE}}$, any $N_{\text{NFE}}\!>\!1$ is a strict efficiency loss with no accuracy upside.

\textbf{Velocity feed-forward.} Removing analytic velocity feed-forward inflates the J7 tracking MAE from $1.33^\circ$ to $2.90^\circ$ (\textbf{$2.18\times$ worse}; $p\!<\!10^{-100}$, $N\!=\!450$ rollouts; Fig.~\ref{fig:ablation}, right). Six of seven joints follow the same trend, confirming the feed-forward's critical role.

\section{Conclusion}
\label{sec:conclusion}

We presented FLASH, a generative visuomotor policy that represents trajectories as \textit{Legendre} polynomial coefficients. Three synergistic mechanisms: sparse temporal sampling, history-anchored flow matching, and analytic velocity feed-forward via polynomial differentiation, jointly enable single-step inference covering a multi-fold extended action horizon, and precise torque-level control. Experiments on five simulated and two real-world tasks confirm state-of-the-art success rates ($\ge$92\%), up to $175\times$ faster inference, $4\times$ faster training convergence and $5\times$--$7\times$ tracking error reductions than prior state-of-the-art policies.

\section{Limitations}
\label{sec:limitations}
Two key design parameters currently lack adaptivity. First, the polynomial degree $K$ must be fixed before training and cannot be adjusted at inference time. The $K\!=\!6$ used throughout may be insufficient for trajectories with sharp, high-frequency components (e.g., contact-rich assembly or dexterous in-hand manipulation). Second, while the execution speed can be freely chosen at inference time by varying $k_{\text{eval}}$, it remains constant throughout a rollout and cannot adapt on the fly to task dynamics (e.g., slowing near a tight insertion while accelerating through free space). A promising future direction is to make both parameters \emph{adaptive}: learning to predict a suitable polynomial degree per segment and to modulate execution speed in a closed loop.

\clearpage
\newpage
\bibliographystyle{assets/plainnat}
\bibliography{paper}

\clearpage
\beginappendix

\section{Cross-horizon continuity and two-tier extension}
\label{appendix:continuity}

This section provides the full derivation of the cross-horizon continuity mechanism summarized in Section~\ref{sec:poly_param}.

\paragraph{First extension: overlap and exact anchors.} The predicted polynomial extends beyond the execution interval of length $T_a$ by $\mu_{\text{pre}}$ steps backward and $\mu_{\text{post}}$ steps forward. This forms an extended evaluation window of total length $H_{\text{poly}}$, representing the actual horizon covered by the predicted polynomial:
\begin{equation}
\label{eq:Hpoly}
H_{\text{poly}} \;=\; \mu_{\text{pre}} + T_a + \mu_{\text{post}}.
\end{equation}
These $\mu_{\text{pre}} + \mu_{\text{post}}$ overlap steps are supervised during training, but discarded before deployment. Within this window, we designate two constraint anchors exactly at the physical boundaries of the execution segment (the front anchor at $t+1$ and the rear anchor at $t+T_a+1$). At these anchors, we explicitly enforce the current polynomial's kinematic states to match the expert trajectory.

This formulation supports arbitrary continuity up to $C^r$ (e.g., $C^3$ for position, velocity, acceleration, and jerk) by appending exact derivatives to the constraints. Stacking these constraints (2 anchors $\times$ \{r + 1\}) into a matrix $\mathbf{A}$ with target $\mathbf{b}$, the closed-form \textit{KKT} solution~\citep{mellinger2011minimum} corrects the coefficients:
\begin{equation}
\label{eq:kkt}
\mathbf{C}^{\star}_{\text{c}}
\;=\;
\mathbf{C}^{\star}
\;-\;
(\mathbf{S}^{\!\top}\mathbf{S})^{-1}\mathbf{A}^{\!\top}
\!\big[\mathbf{A}(\mathbf{S}^{\!\top}\mathbf{S})^{-1}\mathbf{A}^{\!\top}\big]^{-1}
\!\big(\mathbf{A}\mathbf{C}^{\star} - \mathbf{b}\big).
\end{equation}

\paragraph{Second extension: fit padding.} For \textit{Legendre} bases, the functions reach extreme values at the boundaries, making the least-squares solution highly sensitive to target perturbations. To prevent degradation of the fit quality near the anchors, we append $\delta$ padding steps to both sides, expanding the full regression window:
\begin{equation}
\label{eq:fitpad}
H_{\text{poly}} \;=\; \delta + \mu_{\text{pre}} + T_a + \mu_{\text{post}} + \delta.
\end{equation}
These $2\delta$ steps serve purely as a data-preprocessing method. They act as additional regression observation points to pull the constraint anchors into the stable interior of the fitting window, without receiving direct training supervision or being output by the network.

\section{Evaluation setup and details}
\label{appendix:eval_setup}
\label{appendix:eval_details}

\paragraph{Tasks and demonstrations.}
For the simulation experiments, we evaluate FLASH on five tasks: \textit{Close Box} from RLBench~\citep{james2020rlbench}, \textit{Stack Cube} and \textit{Pick Cube} from ManiSkill~\cite{mu2021maniskill} (using the Isaac simulator), as well as \textit{Pick-Place Bowl} and \textit{Open Drawer} from LIBERO~\cite{liu2023libero} (using the MuJoCo simulator). We collected 100 expert demonstrations for \textit{Close Box}, \textit{Pick Cube} and \textit{Stack Cube}, and 40 demonstrations for \textit{Pick-Place Bowl} and \textit{Open Drawer}. All models are trained for 10k optimizer steps and evaluated over 50 rollouts with randomized initial states.

\paragraph{Hyperparameters.}
To ensure a fair comparison, all methods share the same visual encoder (ResNet-18~\citep{he2016deep} with group normalization), observation horizons, batch size, and optimizer schedule; the only varying factors are the algorithmic core and method-specific settings. Through extensive hyperparameter search, we find that A2A-Noise achieves its best performance with a noise scale of $\epsilon=0.02$ and a single-step ($\text{NFE}=1$) inference, which outperforms the original A2A with 6-step inference. We therefore report results with this best-performing configuration. For FLASH and FLASH-G, we use temporal sampling stride $k=4$, Legendre degree $K=6$, and FLASH's unique setting is: a single-step Euler solver, the noise scale injected to the history $\sigma_h=0.5$, \textit{Tikhonov} regularization term $\lambda_h=0.1$, and the weight for consistency loss $\lambda_{\text{cons}}=1.0$.

\section{Success rate analysis}
\label{appendix:sr_analysis}

Table~\ref{tab:sr_main} presents the success rates of all 11 policies across the five simulated tasks under a 10k-step training budget. The empirical results demonstrate a three-tier progression:

\par\noindent\hspace{0.5cm}\hangindent=0.8cm\textbullet\textbf{State-of-the-art performance:} FLASH ranks first on 4 out of 5 tasks, ties for first with ACT on \textit{Pick Cube}, and is the only method to achieve $\ge 92\%$ success rate across all five tasks, requiring only a single sampling step (NFE=1) during inference.
\par\noindent\hspace{0.5cm}\hangindent=0.8cm\textbullet\textbf{Significant improvement over FLASH-G:} FLASH outperforms FLASH-G by an average of 17.6 absolute percentage points (79.2\% $\to$ 96.8\%). Under identical representations, architectures, and hyperparameters, merely replacing the noise prior with the history polynomial yields this substantial gap, quantitatively validating the contribution of the FLASH mechanism detailed in Sec.~\ref{sec:p2p_flow}.
\par\noindent\hspace{0.5cm}\hangindent=0.8cm\textbullet\textbf{Advantage over the strongest baseline:} FLASH maintains a 2\% to 8\% lead (average 4.8\%) over A2A-Noise. While A2A-Noise utilizes an informative prior within a discrete action space, FLASH elevates this prior to the polynomial coefficient space, raising the performance ceiling under the same inference budget.

These advantages are largely attributed to the smoother trajectories from sparse sampling: Outputs are constrained to smooth polynomials. The temporally downsampled $H_{\text{poly}}$ window covers a vastly extended physical horizon (by a factor of $k=4$), drastically mitigating action jitter and unexecutable commands, which is particularly crucial for long-horizon tasks like \textit{Close Box} and \textit{Pick-Place Bowl}. Moreover, position and velocity continuity at chunk junctions in training data are explicitly guaranteed by the KKT-constrained least squares (Eq.~\eqref{eq:kkt}), further alleviating switching jitter between adjacent polynomials.

\section{Training efficiency: remaining tasks}
\label{appendix:training_efficiency}

\begin{figure}[ht]
\centering
\begin{subfigure}[b]{0.49\textwidth}
    \includegraphics[width=\linewidth]{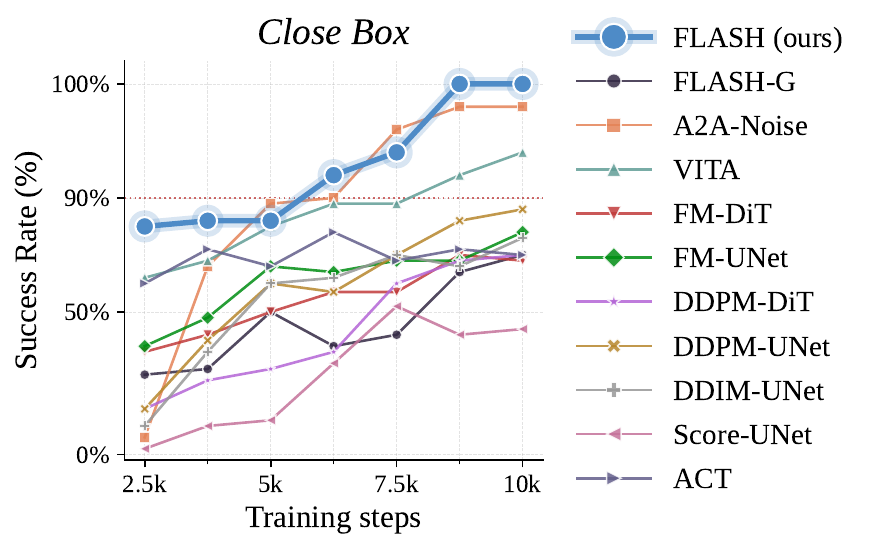}
    \caption{\textit{Close Box}}
\end{subfigure}
\hfill
\begin{subfigure}[b]{0.49\textwidth}
    \includegraphics[width=\linewidth]{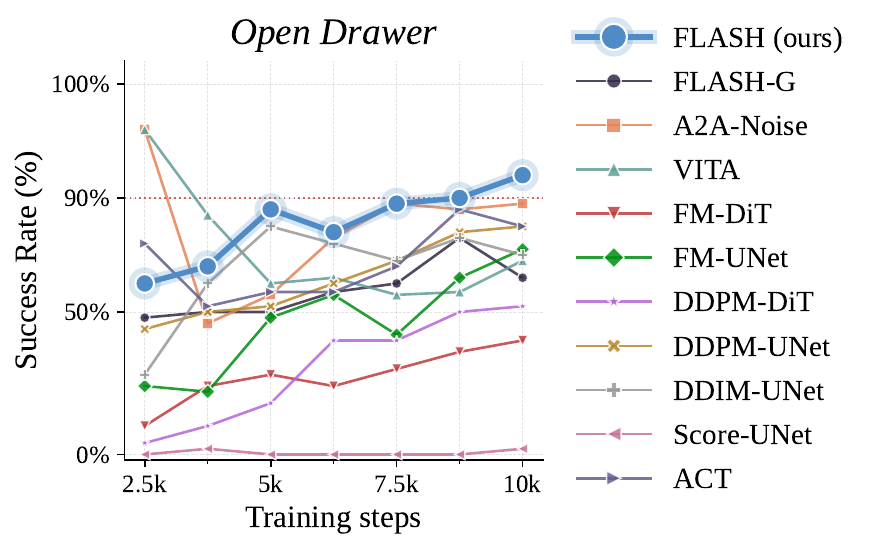}
    \caption{\textit{Open Drawer}}
\end{subfigure}
\caption{\textbf{Training efficiency on remaining tasks.} Same protocol as Fig.~\ref{fig:training_efficiency}.}
\label{fig:training_remaining}
\end{figure}

Fig.~\ref{fig:training_remaining} shows the training curves for the two tasks not included in the main text. On \textit{Close Box}, FLASH reaches 100\% at 10\,k steps while most baselines plateau below 90\%. On \textit{Open Drawer}, FLASH converges to 92\% at 10\,k steps, while strongest competitor at that point (A2A-Noise) lags by 10\,pp.

\section{Inference speed: additional metrics}
\label{appendix:inference_speed}

This section provides the two additional inference speed metrics summarized in Section~\ref{sec:eval_speed}.

\begin{figure}[ht]
\centering
\includegraphics[width=0.55\linewidth, trim=0 25 0 0, clip]{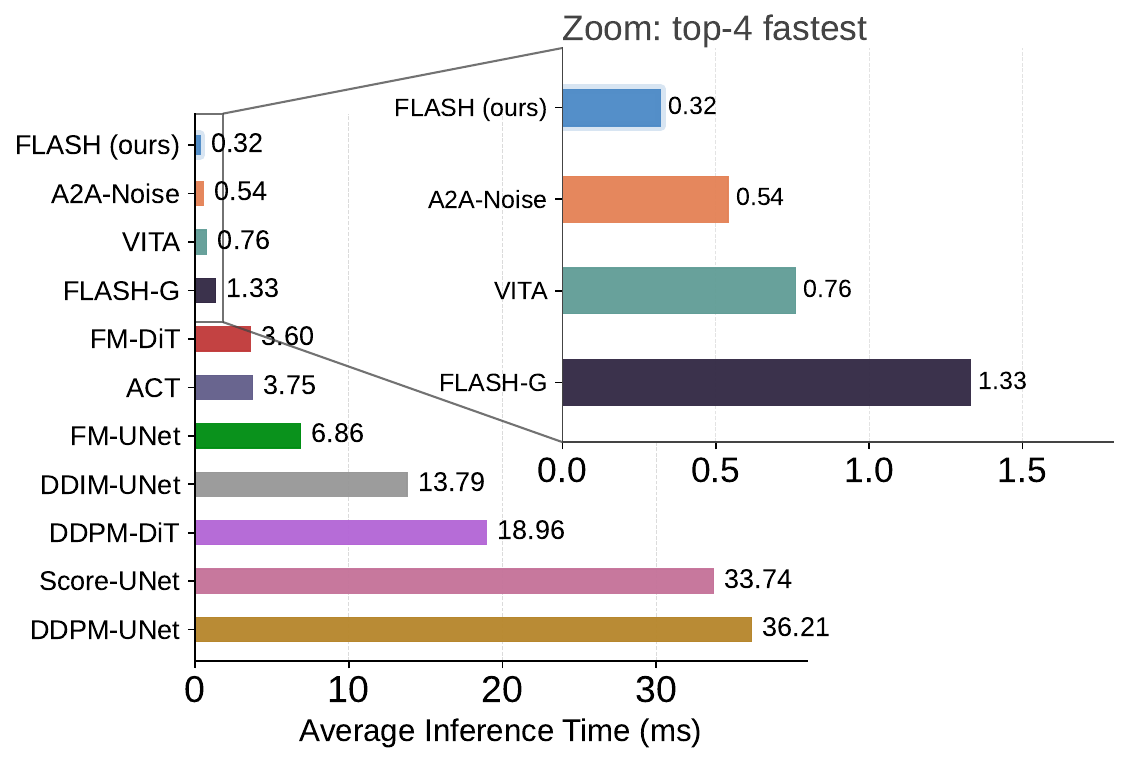}
\caption{Per-call latency (Average Inference Time) on simulated Stack Cube.}
\label{fig:inference_percall}
\end{figure}

\paragraph{Per-call latency (Average Inference Time).}
As shown in Fig.~\ref{fig:inference_percall}, this metric is defined as the amortized inference time per environmental step (total inference time divided by total environmental steps). It fully amortizes chunking and NFE, remains independent of success rates or episode lengths. FLASH ranks first at 0.32~ms:
\par\noindent\hspace{0.5cm}\hangindent=0.8cm\textbullet\ $113\times$ faster than DDPM-UNet (36.21~ms);
\par\noindent\hspace{0.5cm}\hangindent=0.8cm\textbullet\ $1.7\times$ faster than the strongest $N_{\text{NFE}}=1$ baseline, A2A-Noise (0.54~ms);
\par\noindent\hspace{0.5cm}\hangindent=0.8cm\textbullet\ $4.2\times$ faster than FLASH-G (1.33~ms), which shares the same polynomials but starts from noise;
\par\noindent\hspace{0.5cm}\hangindent=0.8cm\textbullet\ $16\times$ faster than the median of the 10 baselines ($\sim$5~ms).

\begin{figure}[ht]
  \centering
  \includegraphics[width=0.55\linewidth]{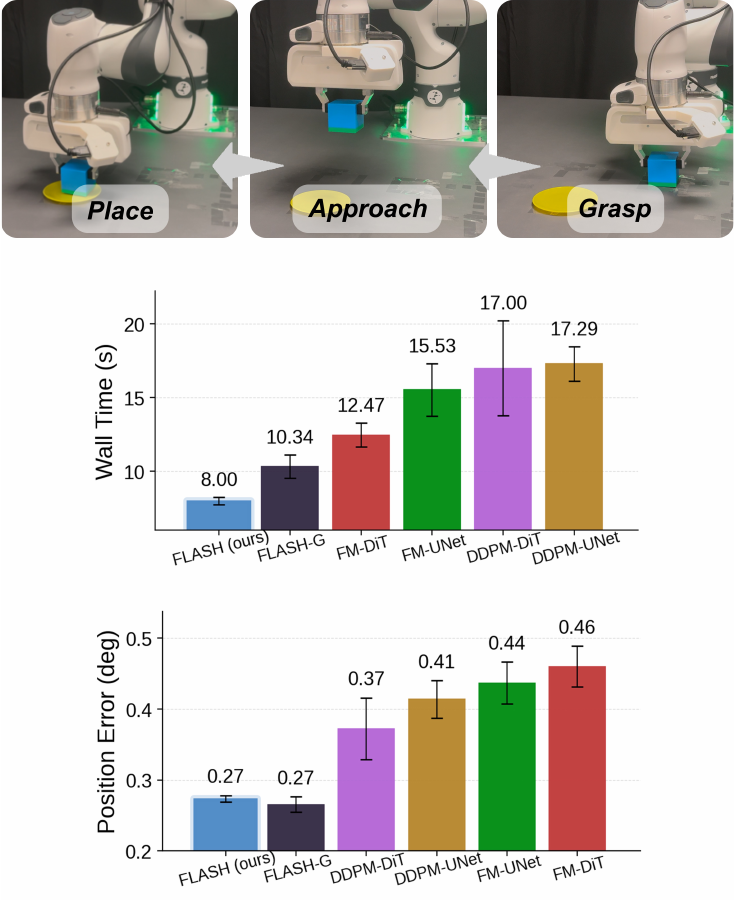}
  \caption{\textbf{Real-world Place Cube.} Top: Execution sequence. Middle: Task completion wall time. Bottom: Tracking error.}
  \label{fig:real_place_cube}
\end{figure}

\paragraph{Real-world task wall time.}
Fig.~\ref{fig:real_place_cube} presents execution snapshots of the real-world \textit{Place Cube} task and compares the wall-clock time across 6 representative policies over 5 rollouts. Real-world latency encompasses the full-pipeline cost of computation, interpolation/communication, and mechanical response, imposing stricter constraints on real-time capabilities than simulation. FLASH completes the task in $8.00 \pm 0.24$~s on average, the fastest among the 6 policies:
\par\noindent\hspace{0.5cm}\hangindent=0.8cm\textbullet\ $2.15\times$ faster than DDPM-UNet (17.29~s) and DDPM-DiT (17.00~s);
\par\noindent\hspace{0.5cm}\hangindent=0.8cm\textbullet\ $1.29\times$ faster than the homologous FLASH-G (10.34~s);
\par\noindent\hspace{0.5cm}\hangindent=0.8cm\textbullet\ Saves $45\%$ of wall-clock time compared to the baseline average of 14.53~s.

\noindent Notably, the standard deviation of FLASH across 5 rollouts is only 0.24~s, an order of magnitude ($13\times$) smaller than that of DDPM-DiT ($\pm3.21$~s), quantitatively reflecting the high execution repeatability of polynomial outputs in the physical world.

\paragraph{Mechanistic breakdown of speed advantages}
\label{appendix:speed_mechanism}

We decompose FLASH's systematic speed advantage into two independent mechanistic contributions, rigorously isolated through pairwise comparisons among FLASH, FLASH-G, and A2A-Noise:

\vspace{0.5em}

\noindent(i) \textit{Sparse sampling training $\rightarrow$ each inference covers $4\times$ horizon.} Both FLASH and FLASH-G utilize the sparse sampling polynomial training strategy (with a temporal stride $k=4$). A single forwardly generated polynomial covers $4\times$ the control duration of discrete action policies, significantly lowering the inference cost. FLASH-G (1.33~ms/call, 159~ms/episode), despite starting from noise, remains faster than almost all baselines except A2A-Noise and VITA, serving as independent empirical evidence for this mechanism.

\vspace{0.5em}
\noindent(ii) \textit{Flow matching initiated from history polynomials.} Although FLASH-G benefits from sparse sampling, its flow still starts from pure \textit{Gaussian} noise and requires multi-step ODE integration. This explains why it is slower than A2A-Noise and VITA. These two baselines both employ an informative prior, indicating that the information content of the starting point intrinsically bounds the speed ceiling. FLASH fuses both mechanisms: sparse sampling + history polynomial starting point, surpassing A2A-Noise by $1.7\times$ at 0.32~ms. FLASH's $4.2\times$ (per call) and $5.1\times$ (per episode) speedups over FLASH-G stem entirely from the difference in the flow's starting point, quantitatively proving the independent contribution of the history-anchored flow mechanism to inference speed.

\section{Trajectory smoothness and tracking error: detailed analysis}
\label{appendix:tracking_details}

This section provides the full analyses summarized in Section~\ref{sec:eval_smooth_tracking}.

\paragraph{Time-series tracking error (full analysis).}
In Fig.~\ref{fig:sim_exp} (middle), we compute the sum of absolute tracking errors across the 7 arm joints (J3 to J9), given by $\sum_{j=3}^{9}|e_j(t)|$. Both the mean and peak errors of FLASH are substantially lower than those of FM-DiT (a mean gap of $4.70\times$ and a peak gap of $3.18\times$). FM-DiT's error curve exhibits spikes at every chunk boundary, a direct consequence of the inter-chunk jitter inherent in discrete action policies. Additionally, FLASH completes the task in just 0.84~s (FM-DiT takes 1.22~s, $1.45\times$ slower), achieving simultaneous compression in both the ``duration'' and ``magnitude'' dimensions of the error.

\paragraph{Cumulative tracking error IAE.}
We also employ the Integral of Absolute Error (IAE), a standard metric in control theory defined as $\mathrm{IAE}=\!\int\!\sum_{j=3}^{9}|e_j(t)|\,dt$ (7 joints / unit: $\text{deg}\cdot\text{s}$), to aggregate the compound cost of ``duration $\times$ magnitude''. For the same \textit{Pick Cube} rollout, FLASH achieves 1.80~$\text{deg}\cdot\text{s}$, while FM-DiT scores 12.23 ($6.80\times$ FLASH).

\paragraph{Real-world control stack (Fig.~\ref{fig:real_place_cube}(bottom)).}
We operate the Franka robot in \textbf{joint torque control} mode, where the host directly computes and sends torque signals to the Franka, rather than relying on Franka's internal dynamic solver. This ensures that tracking errors and practical control frequencies directly reflect the policy's output quality, unmasked by internal engineering smoothing techniques. Notably, due to the inherent properties of continuous polynomials, FLASH and FLASH-G can output signals directly at 1000~Hz, whereas the other 4 baselines must rely on 10$\rightarrow$1000~Hz linear interpolation to perform control. Moreover, FLASH and FLASH-G analytically provide the desired velocity $\dot{q}_d$ via polynomial differentiation (Eq.~\ref{eq:decode}), directly supplying the torque controller's velocity feed-forward term; discrete-action baselines lack this capability and must resort to noisy numerical differentiation, degrading tracking precision. This setup equally applies to the simulated experiments.

We measure the control-level 7-joint average tracking MAE over 6 policies $\times$ 5 rollouts. FLASH (0.274 $\pm$ 0.004 deg) and FLASH-G (0.266 $\pm$ 0.011 deg) perform virtually identically (the 0.008 deg mean difference is smaller than their respective inter-rollout standard deviations, rendering them statistically indistinguishable). DDPM-DiT, DDPM-UNet, FM-UNet, and FM-DiT yield 0.373, 0.414, 0.437, and 0.460 deg, respectively (representing a $1.36\times$ to $1.68\times$ error increase relative to FLASH). The inter-rollout standard deviation of FLASH is only $\pm$0.004 deg, the smallest among all policies and an order of magnitude lower than DDPM-DiT's $\pm$0.043 deg, reflecting the high execution repeatability of polynomial trajectories on real hardware.

\paragraph{High-precision real-world \textit{Insert Cube} task (full analysis).}
We validate FLASH under strict high-precision constraints on the real-world \textit{Insert Cube} task (a precision insertion with millimeter-level tolerance). All models are trained with 10k optimizer steps. The first three subfigures of Fig.~\ref{fig:insert_cube} chronologically illustrate the execution process (grasp $\rightarrow$ approach $\rightarrow$ insert). The fourth subfigure summarizes the real-world success rates of 8 policies using a radar chart: FLASH ranks first with 100\%, leading the other 7 baselines by an average of 47\%. Notably, while DDPM-UNet performs adequately in simulation (84\%), its success rate plummets to only 10\% on the real robot, highlighting the fragility of discrete action diffusion policies under high-precision physical constraints.

Across both simulated (Fig.~\ref{fig:sim_exp}, middle) and real-world results (Fig.~\ref{fig:real_place_cube}~(bottom), Fig.~\ref{fig:insert_cube}), FLASH is consistently superior to non-polynomial baselines. This confirms that control-level tracking precision is jointly improved by three advantages exclusive to polynomial parameterization (the independent contribution of Section~\ref{sec:poly_param}): output trajectory smoothness, native high-frequency control signal generation, and analytic velocity feed-forward for torque control. These three factors are orthogonal to the flow starting point (the FLASH-exclusive mechanism of Sec.~\ref{sec:p2p_flow}).

\section{Post-hoc execution speed modulation}
\label{appendix:speed_mod}

We empirically validate the post-hoc speed modulation property established in Section~\ref{sec:poly_param} on \textit{Stack Cube} using the 10\,k step FLASH checkpoint. We fix the training stride $k_{\text{train}}\!=\!4$ and sweep the evaluation stride $k_{\text{eval}}\!\in\!\{1,\dots,12,16,32\}$, which spans playback speeds from $4\times$ acceleration to $8\times$ deceleration relative to the training pace. Figure~\ref{fig:sim_exp} (right) summarizes the resulting success rates.

FLASH maintains $\ge\!93\%$ success across $k_{\text{eval}}\!\in\!\{3,5,6,7,8\}$, forming a wide high-SR plateau spanning $0.5\times$ to $1.33\times$ playback speed; notably, $k_{\text{eval}}\!\in\!\{5,7\}$ surpass the training value $k_{\text{eval}}\!=\!4$ ($98\%$ vs.\ $96\%$), as a slightly larger $T$ supplies the controller with denser intermediate commands, more than offsetting the mild observation distribution shift. The two failure modes are asymmetric: on the acceleration side, the most extreme $4\times$ speed-up still yields a non-zero $26\%$ success rate; on the deceleration side, the collapse is sharper ($\ge\!2.5\times$ slow-down fails outright), because over-stretched observation intervals leave the robot displaced too far between consecutive inferences.

\section{Ablation study details}
\label{appendix:ablation}

\subsection{Sparse sampling stride $k$}
\label{appendix:abl_stride}

We sweep the temporal stride $k$ defined in Section~\ref{sec:poly_param} from $1$ to $18$ on the simulated \textit{Stack Cube} task with $400$ demonstrations, fixing the \textit{Legendre} degree at $K=6$, and evaluate $12$ checkpoints every $1{,}250$ training steps. Fig.~\ref{fig:ablation} (left) reports the result.

\par\noindent\hspace{0.5cm}\hangindent=0.8cm\textbullet\textbf{Necessity of sparse sampling.} Without sparse sampling ($k=1$) the success rate caps at $24\%$ throughout training; raising $k$ to $4$ lifts it to $96\%$, a $72$\,pp gap that quantitatively justifies the strategy.

\par\noindent\hspace{0.5cm}\hangindent=0.8cm\textbullet\textbf{Bell-shaped optimum.} A sweet spot emerges at $k\!\in\![3,8]$ where every run converges to $\ge 93\%$, with $k=8$ crossing $80\%$ in only $5{,}000$ steps (Fig.~\ref{fig:ablation}, left~(a)). Beyond $k\!\ge\!13$ performance collapses ($52\%$ at $k=15$, $28\%$ at $k=18$). Two factors compound here: (i) the fixed-degree polynomial (Eq.~\ref{eq:ols_fit}) loses expressivity as the physical span $k\!\cdot\!H_{\text{poly}}$ grows; (ii) a single inference covers too much of the task that only $\sim\!2.7$ forward passes per successful episode remain at $k=18$, leaving the policy little chance to revise its plan from fresh visual feedback.

\par\noindent\hspace{0.5cm}\hangindent=0.8cm\textbullet\textbf{Pareto trade-off and practical speed limit.} Per-call latency falls monotonically from $0.97$~ms ($k=1$) to $0.15$~ms ($k=14$), and per-episode total inference time tracks it from $116$~ms to $14.5$~ms. Pushing $k$ from $4$ to $8$ retains $98\%$ success while halving both metrics to $0.21$~ms~/~$20.3$~ms. We retain $k=4$ as the main-paper default, sweet-spot accuracy paired with a $3.7\times$ speed-up over $k=1$, and treat $k=8$ as a more aggressive setting whenever inference budget is the binding constraint.

\subsection{Fit padding and cross-horizon continuity}
\label{appendix:abl_fitpad}

We isolate the contribution of the two-tier extension introduced in Section~\ref{sec:poly_param} on the \textit{Close Box} task, sharing identical training pipelines across the three configurations (last-4-checkpoint average success rate reported in Table~\ref{tab:abl_fitpad}).

\begin{table}[h]
\centering
\small
\setlength{\tabcolsep}{8pt}
\caption{\textbf{Ablation of fit padding and KKT continuity on \textit{Close Box}.}}
\label{tab:abl_fitpad}
\begin{tabular}{lc}
\toprule
\textbf{Configuration} & \textbf{Success Rate (\%)} \\
\midrule
$\delta=0$, no KKT continuity              & 75.0 \\
$\delta=1$, no KKT continuity              & 86.0 \;($+11.0$) \\
$\delta=1$ + KKT continuity (default)      & \textbf{97.5} \;($+11.5$) \\
\bottomrule
\end{tabular}
\end{table}

Fit padding pulls the constraint anchors into the stable interior of the \textit{Legendre} fitting window and yields an $11.0$\,pp gain; stacking the KKT $C^1$ correction (Eq.~\ref{eq:kkt}) on top contributes another $11.5$\,pp, jointly closing a $22.5$\,pp gap to a near-saturated $97.5\%$. Both extensions are therefore necessary for the polynomial pipeline.

\subsection{Number of function evaluations $N_{\text{NFE}}$}
\label{appendix:abl_nfe}

We sweep the inference $N_{\text{NFE}} \in \{1, 2, 4, \dots, 60\}$ on a single \textit{Close Box} checkpoint trained for $2{,}500$ steps; Fig.~\ref{fig:ablation} (middle) reports success rate and per-call latency.

The single-step Euler ($N_{\text{NFE}}=1$) reaches $80\%$ success rate, \emph{outperforming} the multi-step integrations at $N_{\text{NFE}}=2$ ($54\%$), $4$ ($56\%$), and $6$ ($68\%$), and matching the saturation level reached by $N_{\text{NFE}}\!\ge\!10$ (mostly $86\%$ to $94\%$, peaking at $96\%$). This counter-intuitive pattern directly validates the consistency loss $\mathcal{L}_{\text{cons}}$ (Eq.~\ref{eq:loss_cons}): by explicitly supervising the $\tau\!=\!0$ one-shot Euler update at training time, the model is specialized for the $N_{\text{NFE}}=1$ regime, whereas small intermediate $N_{\text{NFE}}$ values lose this specialization without yet accumulating enough multi-step accuracy. Meanwhile, per-call latency rises nearly linearly with $N_{\text{NFE}}$, from $0.32$~ms at $N_{\text{NFE}}=1$ to $3.56$~ms at $N_{\text{NFE}}=60$ ($11\times$), so any $N_{\text{NFE}}\!>\!1$ is a strict efficiency loss with no accuracy upside, justifying our deployment choice of single-step Euler integration.

\subsection{Velocity feed-forward}
\label{appendix:abl_vel}

FLASH derives the desired velocity signal analytically from the polynomial coefficients (Eq.~\ref{eq:decode}), feeding it directly to the torque controller as a feed-forward term. To isolate the contribution of this component, we ablate the velocity feed-forward by sending only position commands (i.e., setting $\hat{\mathbf{v}} = \mathbf{0}$) while keeping all other components unchanged: the polynomial representation, sparse sampling, FLASH flow matching, and KKT continuity correction.

\paragraph{Experimental setup.}
We evaluate on the same 9 checkpoints (spanning 3 tasks $\times$ 3 training seeds) used in the preceding ablation, with 50 rollouts per checkpoint ($N = 450$ total). For each rollout, we record the mean absolute error (MAE, in degrees) between the commanded and actual joint positions across all timesteps, then compare the \textit{with-velocity} and \textit{without-velocity} conditions in a pairwise fashion.

\paragraph{Results (Fig.~\ref{fig:ablation}, right).}
Removing the analytic velocity feed-forward inflates the J7 (wrist) position-tracking MAE from $1.33^\circ$ to $2.90^\circ$, a $\mathbf{2.18\times}$ degradation. We report three complementary statistical measures to characterize this effect:

\begin{itemize}[leftmargin=0.5cm]
    \item \textbf{Paired $t$-test ($p < 10^{-100}$):} Because each rollout is executed under both conditions on the same checkpoint, we use a \textit{paired} (dependent-samples) $t$-test rather than an independent two-sample test. The resulting $p$-value is astronomically small, indicating that the observed MAE increase is far beyond what random variation could explain.
    \item \textbf{Cohen's $d = 1.78$:} While $p$-values indicate \textit{whether} an effect exists, Cohen's $d$ quantifies \textit{how large} it is by expressing the mean difference in units of the pooled standard deviation. By conventional benchmarks~\citep{cohen2013statistical}, $d \ge 0.8$ is already considered a ``large'' effect; the observed $d = 1.78$ is more than twice that threshold, confirming that the velocity feed-forward produces a practically significant (not merely statistically significant) improvement.
    \item \textbf{Sample size ($N = 450$):} Reporting the sample size is essential for interpreting both the $p$-value and Cohen's $d$. With 450 paired observations, the test has ample statistical power to detect even moderate effects; the extremely large $d$ and vanishingly small $p$ together rule out both Type~I and Type~II error concerns.
\end{itemize}

Six of seven arm joints exhibit the same pattern: the position-only controller consistently produces larger tracking errors, with the degradation most pronounced in the distal joints (J6, J7) where inertia is lower and high-frequency feed-forward compensation matters most. This confirms that the analytic velocity feed-forward is not merely a theoretical convenience but a critical component for precision torque control.

\end{document}